\documentclass[sigconf]{acmart}
\AtBeginDocument{%
  }

\setcopyright{acmlicensed}
\copyrightyear{2025}
\acmConference[CIKM '25]{Proceedings of the 34th ACM International Conference on Information and Knowledge Management}{November 10--14, 2025}{Seoul, Republic of Korea}
\acmBooktitle{Proceedings of the 34th ACM International Conference on Information and Knowledge Management (CIKM '25), November 10--14, 2025, Seoul, Republic of Korea}

\usepackage[para]{footmisc}

\begin{document}

\title{Explain and Monitor Deep Learning Models for Computer Vision using Obz AI}

\author{Neo Christopher Chung}
\affiliation{%
  \institution{Institute of Informatics, University of Warsaw}
  \institution{Alethia XAI Sp. z o.o.}
  \city{Warsaw}
  \country{Poland}
}
\email{nchchung@gmail.com}

\author{Jakub Binda}
\affiliation{%
  \institution{Institute of Informatics, University of Warsaw}
  \institution{Alethia XAI Sp. z o.o.}
  \city{Warsaw}
  \country{Poland}}
\email{j.binda@student.uw.edu.pl }

\renewcommand{\shortauthors}{Chung and Binda}

\begin{abstract}
Deep learning has transformed computer vision (CV), achieving outstanding performance in classification, segmentation, and related tasks. Such AI-based CV systems are becoming prevalent, with applications spanning from medical imaging to surveillance. State of the art models such as convolutional neural networks (CNNs) and vision transformers (ViTs) are often regarded as ``black boxes,'' offering limited transparency into their decision-making processes. Despite a recent advancement in explainable AI (XAI), explainability remains underutilized in practical CV deployments. A primary obstacle is the absence of integrated software solutions that connect XAI techniques with robust knowledge management and monitoring frameworks.

To close this gap, we have developed \emph{Obz AI}, a comprehensive software ecosystem designed to facilitate state-of-the-art explainability and observability for vision AI systems. Obz AI provides a seamless integration pipeline, from a Python client library to a full-stack analytics dashboard. With Obz AI, a machine learning engineer can easily incorporate advanced XAI methodologies, extract and analyze features for outlier detection, and continuously monitor AI models in real time. By making the decision-making mechanisms of deep models interpretable, Obz AI promotes observability and responsible deployment of computer vision systems.
\indent Obz AI: \url{https://obz.ai} 
\end{abstract}

%

\begin{teaserfigure}
  \includegraphics[width=\textwidth]{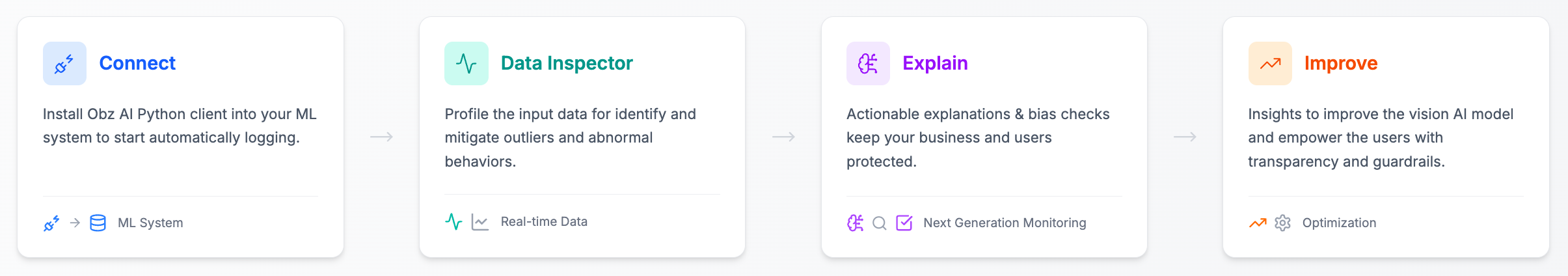}
  \caption{Overview of Obz AI consisted of the Python library, the database/backend, and the front-end dashboard.}
  \Description{Obz AI -- consisted of the Python client library, the database, the back-end server, and the front-end dashboard -- enable advanced monitoring of AI systems with computer vision applications.}
  \label{fig:overview}
\end{teaserfigure}


\maketitle

\section{Introduction}

Deep learning has drastically improved a wide range of computer vision (CV) tasks such as image classification and segmentation \citep{vaswani2017attention, dosovitskiy2020image, Wu2020}. These models are now deployed widely in high-stakes domains from medical imaging and diagnostics to surveillance and autonomous driving.  However, despite state-of-the-art performance, popular vision models such as vision transformers (ViT) and convolutional neural networks (CNN) remain ``black boxes'' whose decision making processes are opaque.  Understanding why a model makes a given prediction is critical for trust and safety, as uninterpretable behaviors hinder fairness, transparency and accountability in real-world vision systems.

To address this, the field of explainable AI (XAI) has produced many methods tailored to CV models.  Early work visualized CNN internals and computed saliency maps by backpropagating gradients to input pixels \cite{simonyan2013deep}.  Guided backpropagation \cite{springenberg2015}, SmoothGrad \cite{smilkov2017smoothgrad} and similar techniques compute per-pixel importance scores via gradients.  Grad-CAM \cite{selvaraju2017grad} uses the gradients with respect to the last layer to produce a coarse heatmap of the image regions most relevant to a class prediction. Due to the self-attention mechanisms, explanation for ViT has mostly relied on attention maps \cite{dosovitskiy2020image}. 
Building upon attention maps, advanced XAI methods for ViT attempt to better integrate information across multiple Transformer blocks \cite{abnar2020quantifying} or to combine them with gradients \cite{brocki2024cdam}. Collectively, these XAI methods offer post-hoc explanations for CV models.

Despite the development of XAI methods, they are rarely integrated into end-to-end vision AI pipelines.  Typical MLOps setups focus on performance metrics and do not incorporate explainability outputs. As a result, practitioners lack unified tools to log, manage and analyze explanations alongside inputs and outputs.  Without such integration, it is hard to systematically monitor a vision model’s reasoning or detect anomalous behaviors in production.

To bridge this gap, we introduce \emph{Obz AI}, a software ecosystem that seamlessly combines state-of-the-art explainability with image feature extraction and outlier detection for vision models (\autoref{fig:overview}). Obz AI provides a full-stack pipeline from a Python client library to an interactive analytics dashboard, enabling AI engineers to:

\begin{itemize}
\item Apply XAI algorithms to a CV model in production.
\item Extract and inspect images to identify outliers and anomalies.
\item Track model predictions and explanations in a dashboard.
\item Store and query explanations alongside metadata for audits.
\end{itemize}

By making the internal decision mechanisms of CV systems more transparent, Obz AI enables explainability and observability, helping users to verify and improve their systems.

\section{Related Works}


Modern ML observability platforms offer end-to-end monitoring of model inputs, outputs, and performance. WhyLabs\footnote{\url{https://whylabs.ai}} offers real-time feature and label drift detection and bias monitoring across production ML pipelines. Evidently\footnote{\url{https://evidentlyai.com}} provides metrics for data/model drift and quality checks, with a central dashboard to track all ML models and datasets over time. However, they focus on tabular and textual data.

Fiddler AI\footnote{\url{https://fiddler.ai}} and Arthur\footnote{\url{https://arthur.ai}} introduced support for vision models that track performances and detect outliers. While they offer limited XAI functionalities, their support for CV models is lacking. On the other hand, Roboflow\footnote{\url{https://roboflow.com}} is tailored to developing CV models, including curating and creating datasets and deploying to production. Currently, Roboflow does not provide XAI functionalities.

There are Python libraries that implements different XAI methods. Dalex is a Python library to provide a model-agnostic interface to XAI methods for tabular data \cite{dalex}. DARPA (Defense Advanced Research Projects Agency) funded development of the explainable AI toolkit (XAITK) \cite{XAITK}. \texttt{saliency (deprecated)} from Google \footnote{\url{https://github.com/PAIR-code/saliency}} and \texttt{Captum} from Meta \footnote{\url{https://captum.ai}} implement a number of saliency maps and other XAI methods, highlighting methods developed in-house. To leverage well-validated methodologies, Obz AI exposes functionalities of \texttt{Captum} to compute different saliency maps, in addition to our own XAI implementations.

\section{System Architecture}

Obz AI is developed as a modular platform to enable explainable monitoring for computer vision (CV) models in production environments. The system is realized as a service-oriented architecture, combining local Python/PyTorch utilities for data inspection and explainability with centralized infrastructure for data storage, orchestration, and visualization. Obz AI is developed from scratch to ensure an easy migration from a cloud to an on-premise server for privacy and compliance. Here, we describe the technical components and their integration, focusing on how Obz AI supports comprehensive model and data auditing.

\subsection{Python/PyTorch Library}

\begin{figure}[t]
  \centering
    \includegraphics[width=.5\textwidth]{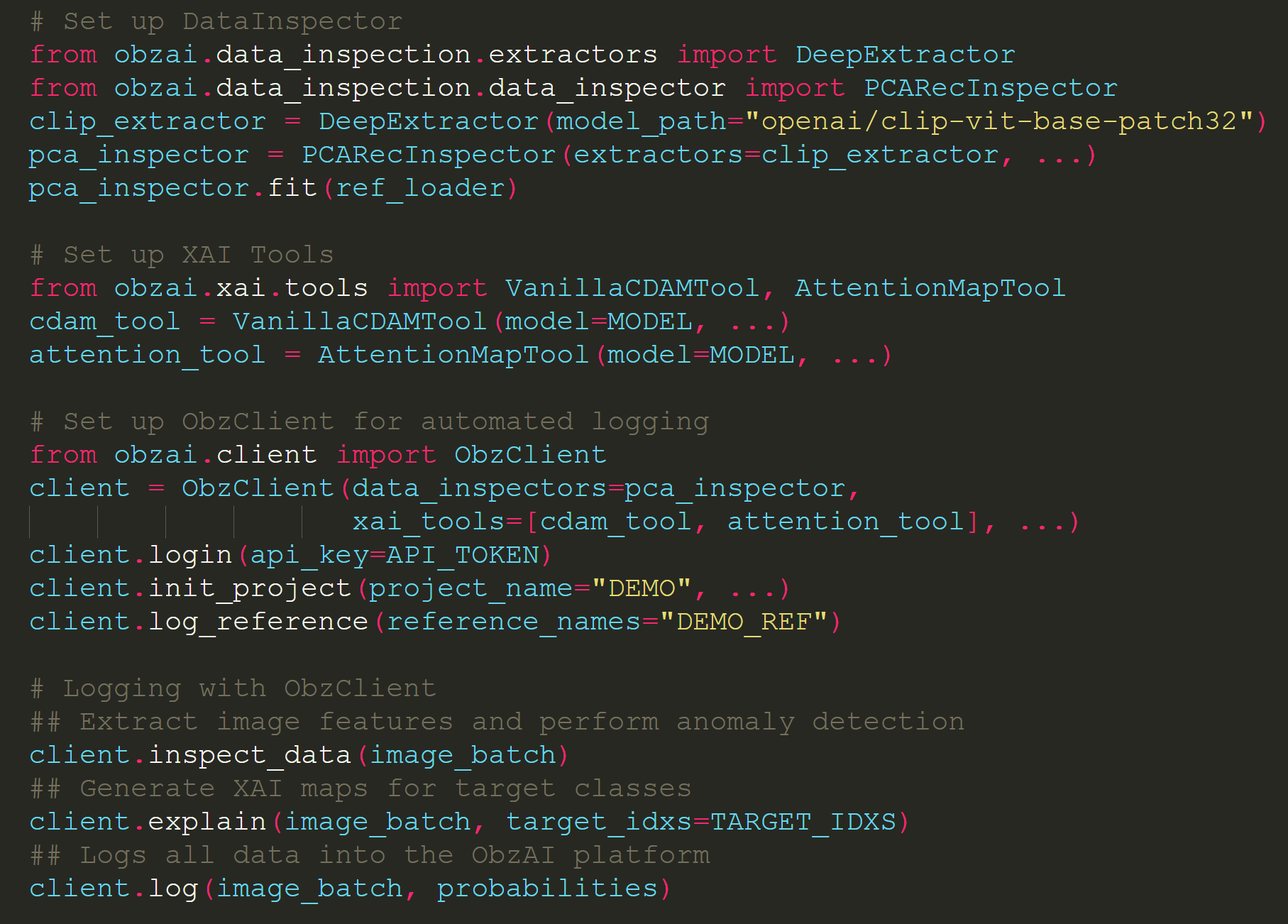}
    \caption{Core functionalities of Obz AI Python Package}
    \Description{Core functionalities of Obz AI Python Package.}
    \label{fig:ObzAI-Python}
\end{figure}

At the core of the system is our open source Python library\footnote{\url{https://pypi.org/project/obzai}}, intended to be installed alongside a user’s AI model. Developed with PyTorch, it directly interfaces with model pipelines and dataset loaders. A illustrative use case is shown in \autoref{fig:ObzAI-Python}. The Python library consist on three primary modules: 

\begin{itemize}
      \item \textbf{XAI} includes explainable AI (XAI) methods and evaluation tools,
      \item \textbf{Data Inspector} characterizes the input data that can be utilized in anomaly detection,
      \item \textbf{Obz Client} collects and saves activities of the AI model on a local and/or cloud server.
\end{itemize}

The XAI module is based on a general-purpose \texttt{XAITool} class facilitating integration with different explainability algorithms. Each XAI tool is designed to process batches of images, returning corresponding importance scores in the same resolution as an input image (so-called an attribution map or an XAI heatmap). For the vision transformer (ViT), several varieties of attention maps are implemented \cite{dosovitskiy2020image}, including CDAM allowing explanations to be targeted at specific model outputs \cite{brocki2024cdam}. The XAI module also inherit explainability methods from the \texttt{Captum} library, which includes SmoothGrad \citep{smilkov2017smoothgrad}, Integrated Gradients \citep{sundararajan2017axiomatic}, and other methods suitable for CNNs.

Evaluation of the explanations is handled by a dedicated submodule, following the abstraction provided by the \texttt{EvalTool} base class. Quantitative assessment metrics include \emph{Fidelity}, which examines the alignment of explanation with the model's predictive behavior based on perturbation curves \citep{brocki2023perturbationartifacts, brocki2023feature}, and \emph{Compactness}, which reflects the potential human interpretability of the maps \citep{brocki2024cdam}.

The data inspection workflow is similarly modular. Feature extraction from input images is abstracted through \texttt{Extractor} classes, with each subclass implementing a specific extraction technique. Inspired by radiomic features \cite{pyradiomics}, we implement first-order feature statistics (e.g., mean, variance, etc) which are then modeled by a Gaussian Mixture Model via the \texttt{DataInspector} class. For advanced spatial features, we use embeddings from visual-language models (VLM), such as CLIP \cite{radford2021clip}, which converts an image into a semantically rich feature vector. Using any \texttt{Auto Model} from the Hugging Face repository, a user can choose a VLM suitable for their application domain. Embeddings are modeled by principal component analysis (PCA) such that a reconstruction loss based on $r$ PCs is used to identify outliers \cite{kennedy2025-outlier}. The outlier detection works by applying the reference model to new images. Application of this data inspection workflow in preclinical imaging is detailed in \citet{binda2025outlier}.

All information -- inputs, raw predictions, XAI maps, features, outlier flags, and others -- are consolidated by the \texttt{ObzClient} class. The client manages user authentication and project scoping on initialization, driven by API tokens and project metadata. Logging methods support flexible packaging of all inference-time results.

\subsection{Data Management and Storage}

Data generated by the monitoring workflows is stored using a combination of relational SQL storage and S3-compatible object storage. This hybrid approach ensures scalable management of both structured metadata (e.g., logs, project definitions, user profiles) and unstructured large files (e.g., raw images and XAI attribution maps). For a centralized cloud solution, we utilize \texttt{PostgreSQL} offered by Supabase.

The SQL database schema organizes data into: 

\begin{itemize}
      \item \textbf{Users} - Standard user records, including credentials.
      \item \textbf{API Tokens} - API Tokens and relate them with users.
      \item \textbf{Projects} - Metadata including the project name, ML task mode, and others.
      \item \textbf{Ref Features} - Logged reference features related with a particular project.
      \item \textbf{Logs} - ID, prediction, input image, XAI maps keys, extracted image features, and others
\end{itemize}

Each project is assigned its own private bucket, ensuring that users’ data is safe and isolated from that of others. The schema supports logging and retrospective analysis at both the project and user level, and is designed for extensibility to evolving ML monitoring requirements.

\subsection{Backend Components}

A Python-based backend, implemented in \texttt{FastAPI}, serves as the central orchestrator for system interactions. The backend exposes authenticated endpoints to the Python client for log ingestion, reference feature upload, and project management. Upon receipt of a log, the backend coordinates storage of metadata in SQL and direct upload of data artifacts to the object storage. It is also responsible for the validation and resolution of API tokens, project uniqueness, and linkage of log records to user and project contexts.

In addition to handling data ingress from the client, the backend provides read APIs to the frontend dashboard, exposing all stored monitoring artifacts to facilitate interactive exploration, visualization, and administrative management. These APIs enable project switching, log inspection, and user account functionality in a secure and scalable manner.

\subsection{Frontend Dashboard}

A web dashboard is implemented in \texttt{TypeScript} and \texttt{React}. It acts as the visible interface for both developers and project managers, aggregating logged data into actionable intelligence. The dashboard enables users to view high-level summaries over time, inspect data and model behaviors, and manage associated resources (projects, API tokens, and users).

From a user experience standpoint, the dashboard is tailored for CV systems. Panels dedicated to data inspection, explainability, and project management allow users to trace an individual log from input image to extracted features, outlier status and generated explanations. From the web dashboard, a user can delete the data or logs. Project switching is integrated throughout the interface, reflecting the multi-project design inherent to large-scale ML operations.

\begin{figure}[t]
  \centering
    \includegraphics[width=.5\textwidth]{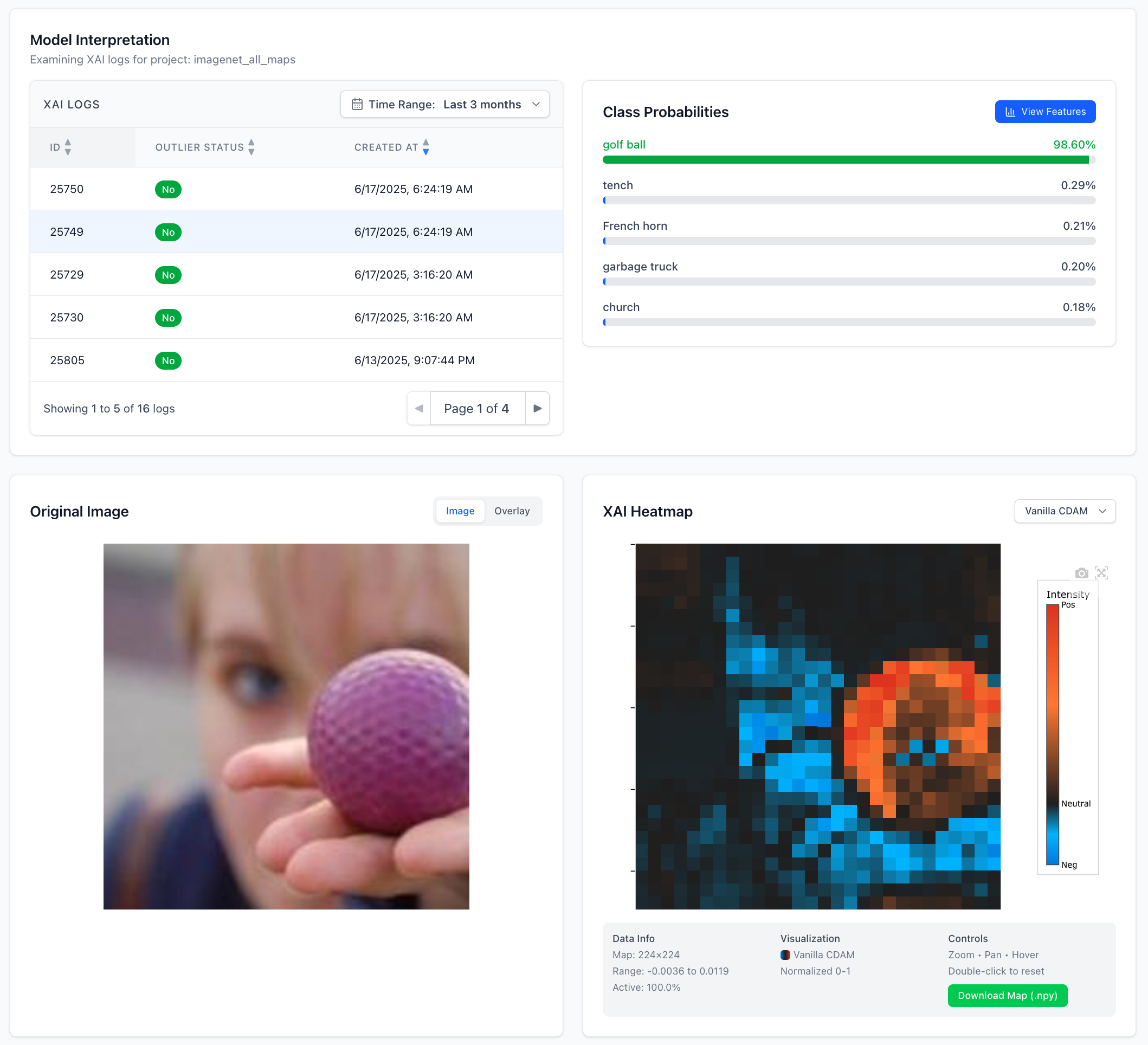}
    \caption{XAI Visualization in Obz AI from the the ImageNet application}
    \Description{XAI Visualization in Obz AI from the the ImageNet application.}
    \label{fig:XAI-ImageNet}
\end{figure}

\section{Applications}

We demonstrate the use of Obz AI functionalities through two example models. Configuration of the Obz AI can be readily changed to accommodate different outlier detection algorithms, VLMs, XAI methods, and ML use cases.

\subsection{Natural Images from The ImageNet}

ImageNet-S50 is a collection of natural images that was initially developed to benchmark unsupervised segmentation models~\cite{gao_large-scale_2023} and been utilized for evaluating XAI methods~\cite{szczepankiewicz_ground_2023}. The dataset covers 50 distinct object categories. This dataset reflects a wide variety of natural images that may be encountered in surveillance, vehicle, and other cameras.

We used a pre-trained multiclass classification ViT model trained on the ImageNet \cite{brocki2024cdam}. In this application, the Data Inspector is configured for the PCA Detector based on CLIP embeddings which are originally trained on a pair of natural images and text descriptors \cite{radford2021clip}. For XAI, we configured the Obz AI to compute Attention Map \cite{dosovitskiy2020image}, CDAM \cite{brocki2024cdam}, and Saliency Map \cite{simonyan2013deep}. After setting up the Obz AI logging pipeline, we can monitor the data inspector, model explanations, and other functionalities on the web app.

\autoref{fig:XAI-ImageNet} show the XAI viewpoint where the original image and XAI heatmap are shown side-by-side. As is common in computer vision applications, we selected the predicted class with the highest probability, namely ``golf ball'' ($98.6\%$). Red and blue indicate to positive and negative importance scores, whereas black represents neutrality ($\sim$0). In this example, pixels inside the golf ball are predominantly associated with positive scores that increase the prediction, whereas pixels outside the golf ball are either slightly negative or neutral. The importance scores in the XAI heatmap can be explored on the web app, or be downloaded as a Numpy array. Selecting ``Overlay'' displays the XAI heatmap on the top of the target image, with a control for transparency. Note that Obz AI provides the flexibility to compute and store XAI not only for the predicted class but also for multiple target classes.

\subsection{Medical Images from LIDC-IDRI}

\begin{figure}[t]
  \centering
    \includegraphics[width=.5\textwidth]{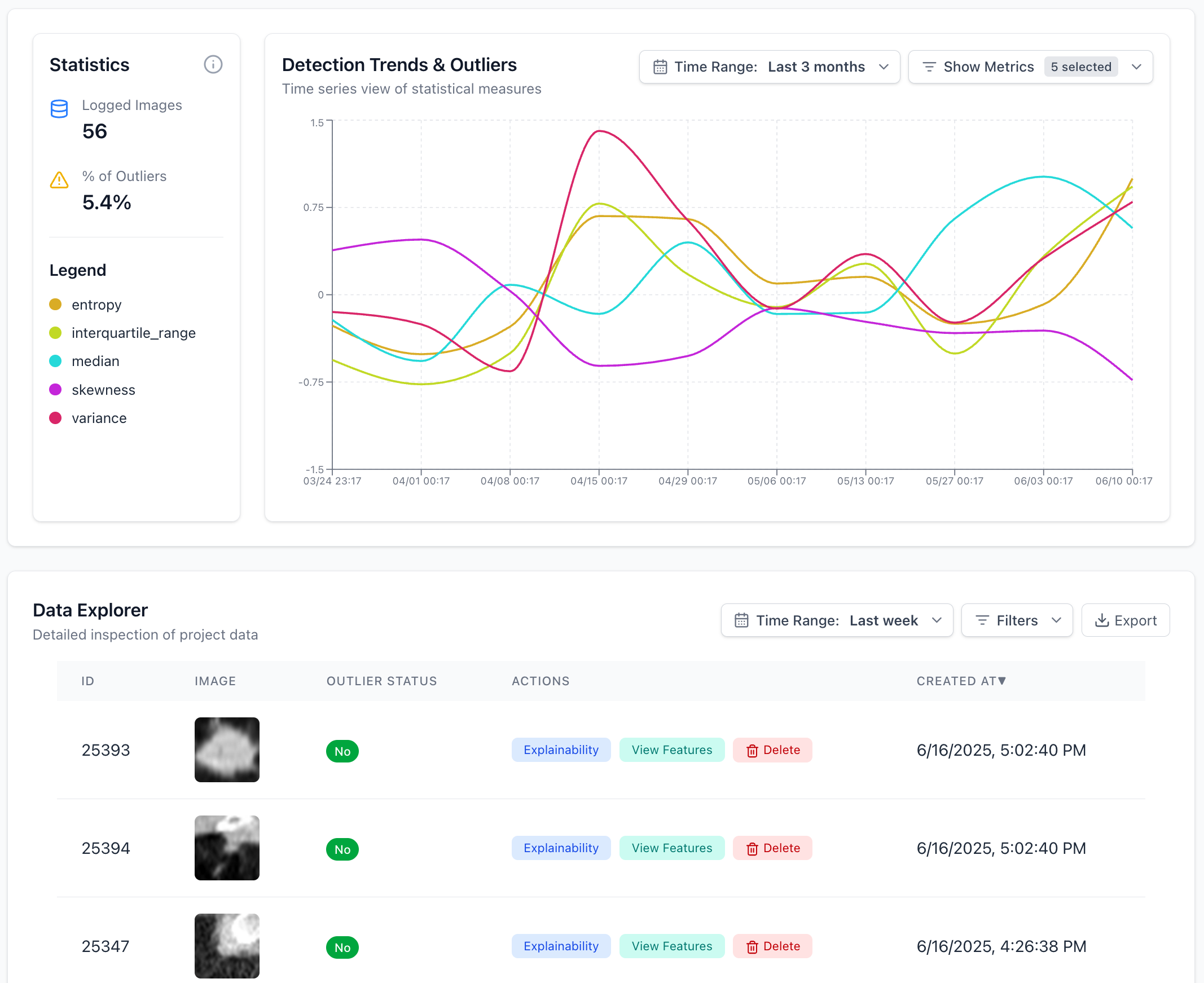}
    \caption{Data Inspector in Obz AI from the LIDC Lung Cancer Application}
    \Description{Data Inspector in Obz AI from the LIDC Lung Cancer Application.}
    \label{fig:DataInspector-LIDC}
\end{figure}

The LIDC-IDRI (Lung Image Database Consortium and Image Database Resource Initiative) dataset comprises chest computed tomography (CT) scans that have been independently annotated by up to four radiologists~\cite{armato2011lung}. For each scan, the radiologists provided nodule segmentation masks as well as malignancy ratings on a five-point scale. The dataset was sourced from the Cancer Imaging Archive (TCIA) \cite{clark2013tcia} and underwent preprocessing as described by~\cite{brocki2023conrad}. We used this dataset as a use case in preclinical or medical applications.

The final dataset included 443 benign and 411 malignant lung nodules CT scans, which are split into training (80\%) and testing (20\%) sets stratified. A binary classifier based on a DINO backbone is trained on the LIDC-IDRI data, predicting malignancy. We fit a Gaussian Mixture Model (GMM) on 16 First Order Features (FOFs).

\autoref{fig:DataInspector-LIDC} shows the Data Inspector viewpoint. At the top, selected features are plotted over time. ``Time Range'' can be modified to show more or less samples recorded. Within a chosen range, we display the total number of samples and the number of outliers. Substantial deviation from a typical distribution in the training set, as modeled by a GMM, would indicate an outlier. While there are 16 FOFs used in GMM, the selected 5 features are visualized for clarity. Additional features can be displayed by selecting within ``Show Metrics''.

In the bottom panel of \autoref{fig:DataInspector-LIDC}, details of an individual sample can be further examined by clicking on ``View Features''. The FOFs of the selected sample is compared to the distributions of FOFs in the reference dataset. Selecting ``Explainability'' opens up an XAI view point for the chosen sample, similar to \autoref{fig:XAI-ImageNet}. Class probabilities are shown, alongside available XAI heat maps (e.g., CDAM, Attention Map, Saliency Maps). Finally, within ``Data Explorer'', usrs have the option display and export a subset of samples as a CSV file for further analysis.

\section{Conclusion and Future Work}

Obz AI assists ML engineers and data scientists in explaining and monitoring computer vision (CV) systems. To the best of our knowledge, this is the first integrated platform tailored to deep learning models for vision tasks. Our open source Python package enables a wide range of outlier detection and explainability methods for CNN and ViT, while the accompanying web app enables seamless visualization and management of logs. 

In this work, we focused on developing Obz AI for classification tasks. We are planning to extend the platform to support image synthesis, segmentation, and additional CV tasks. Thus far, our efforts have concentrated on monitoring and explaining ViT and CNN models through implementations of modern XAI methods. An important direction for future development is the integration of more conventional ML models and XAI methods to broaden the applicability of the platform.

With Obz AI, we anticipate that a wide range of CV systems will be able to seamlessly integrate XAI and outlier detection methodologies. Although large language models (LLMs) currently dominate the research and development in observability and monitoring solutions, we foresee growing opportunities for solutions tailored to visual domains, which are becoming increasingly critical in medicine, autonomous driving, and other industries.

\begin{acks}
Ruslan Bayramov developed the web app. Lennart Brocki provided technical advice. We acknowledge George Dyer and Hongkyou Chung for their compliance and legal consultations.
\end{acks}

\section*{GenAI Usage Disclosure}
In accordance with ACM’s policy on the use of generative AI, we affirm that GitHub Copilot was used in programming and ChatGPT and DeepL were used to revise and edit the manuscript. Authors have fully reviewed all of AI usage and are fully accountable for the content.

\bibliographystyle{ACM-Reference-Format}
\bibliography{citations}



\end{document}